%% file: latex/acl_latex.tex
\documentclass[11pt]{article}

\usepackage[final]{acl}
\usepackage[para]{footmisc}
\usepackage{stfloats}
\usepackage{times}
\usepackage{latexsym}
\usepackage{xcolor}
\input{math_commands.tex}

\usepackage{hyperref}
\usepackage{url}

\usepackage{booktabs}    
\usepackage{multirow} 
\usepackage{enumitem}
\usepackage{amsmath}      
\usepackage{amssymb}      
\usepackage{mathtools}    
\usepackage{xcolor}
\usepackage{times}  
\usepackage{helvet}  
\usepackage{pifont}
\usepackage{amsmath}
\usepackage{amssymb}
\usepackage{subfig}
\usepackage{subcaption}
\usepackage[table]{xcolor}

\usepackage{graphicx}
\usepackage{algorithm}

\usepackage{enumitem}
\usepackage{float}
\usepackage{amsmath}
\usepackage{amssymb}
\usepackage{amsfonts}
\usepackage{algorithmic}
\usepackage{subfig}
\usepackage{amsmath}      
\usepackage{amssymb}      
\usepackage{mathtools}    
\usepackage{dsfont}       
\usepackage{bbm} 

\usepackage[T1]{fontenc}

\usepackage[utf8]{inputenc}

\usepackage{microtype}

\usepackage{inconsolata}

\usepackage{graphicx}

\title{BCL: Bayesian In-Context Learning Framework for Information Extraction}

\author{\normalfont
Haoliang Liu\textsuperscript{1}\thanks{Equal contribution.} \quad
Chengkun Cai\textsuperscript{2}\footnotemark[1] \quad
Xu Zhao\textsuperscript{3}\footnotemark[1] \quad
Han Zhu\textsuperscript{4} \quad
Shizhou Huang\textsuperscript{5}\thanks{Important contribution.} \\
Xinglin Zhang\textsuperscript{6} \quad
Tao Chen\textsuperscript{7} \quad
Jenq-Neng Hwang\textsuperscript{8} \quad
Zhang Huaping\textsuperscript{9}\quad
Lei Li\textsuperscript{9} \thanks{\parbox[t]{0.9\linewidth}{Corresponding authors: \texttt{lilei@bit.edu.cn}}}
}

\begin{document}
\maketitle

\begingroup
\renewcommand\thefootnote{}
\makeatletter
\renewcommand\@makefntext[1]{\noindent #1}
\makeatother
\footnotetext{
\textsuperscript{1}HiThink Research \quad
\textsuperscript{2}University College London \\
\textsuperscript{3}University of Edinburgh \\
\textsuperscript{4}The Hong Kong University of Science and Technology \\
\textsuperscript{5}East China Normal University \\
\textsuperscript{6}Shanghai Medical Image Insights \\
\textsuperscript{7}University of Waterloo \quad
\textsuperscript{8}University of Washington \\
\textsuperscript{9}Beijing Institute of Technology
}
\endgroup

\begin{abstract}

Existing information extraction (IE) tasks increasingly adopt in-context learning (ICL) with large language models. However, current approaches either show inconsistent performance across model scales or lack systematic optimization and generalizability. Building on this, we propose BCL (Bayesian In-Context Learning Framework for Information
Extraction), the first optimization framework that uses particle filtering with Bayesian updates to systematically refine label representations across IE tasks. Through four steps—initialization, observation, weight update, and resampling, BCL generalizes to both sequence labeling and relation classification paradigms. Extensive experiments demonstrate substantial and consistent improvements over existing approaches.

\end{abstract}

\section{Introduction}

Recent IE tasks rely on in-context learning (ICL), where large language models (LLMs)~\citep{brown2020language} are guided by contextual information. Recent approaches can be broadly categorized into task transfer approaches that reformulate information extraction (IE) as auxiliary tasks (e.g., ChatIE~\citep{wei2023chatie}, CodeIE~\citep{li2023codeie}) and guideline-based approaches that provide explicit annotation guidelines (e.g., GuideNER~\citep{huang2025guidener}).

\begin{figure}[t]
\centering
\includegraphics[width=1.05\linewidth]{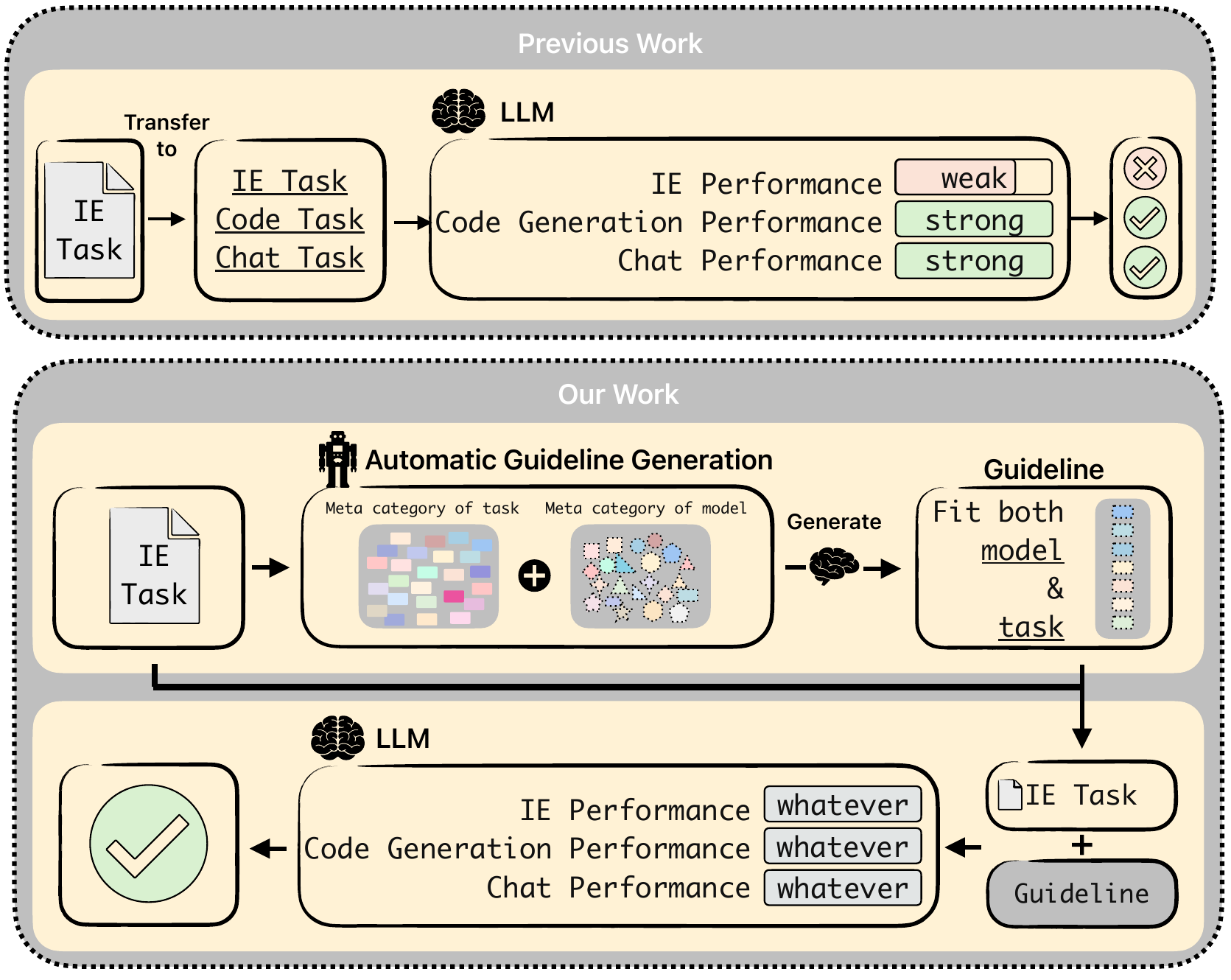}
\caption{\textbf{Comparison of previous approaches and our work.} 
\textbf{Top:} Previous methods convert IE tasks to leverage models' stronger 
code or chat capabilities, making performance dependent on model-specific strengths. 
\textbf{Bottom:} Our approach directly improves IE performance via automatically 
generated semantic patterns, regardless of models' relative strengths across task types.}
\label{fig:motivation}
\end{figure}

However, existing approaches face practical limitations. As illustrated in 
Figure~\ref{fig:motivation} (top), task transfer methods show inconsistent 
performance across model scales. While potentially effective on ultra-large 
commercial models, they often underperform direct IE prompting on 
smaller models. ChatIE underperforms one-shot prompting by a substantial margin on NER tasks, and CodeIE fails on RE tasks with near-zero micro-F1. This inconsistency makes deployment challenging when using lightweight models, which are common in practical settings due to computational constraints.

\begin{figure}[!htbp]
\centering
\includegraphics[width=1.05\linewidth]{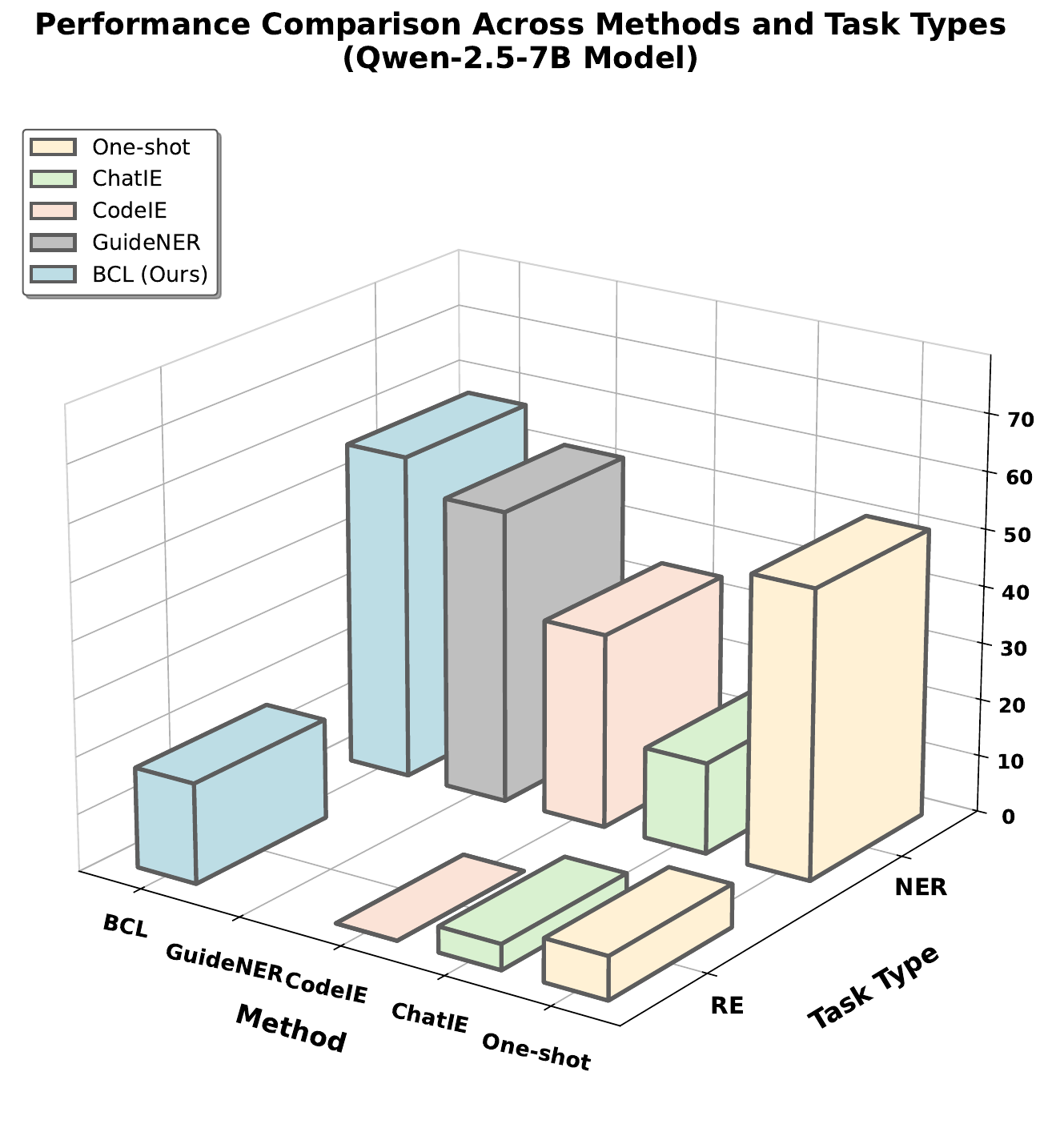}
\caption{Performance comparison of different methods on Qwen-2.5-7B across NER and RE tasks. The Y-axis represents F1 score (\%). BCL demonstrates consistent superiority over baseline methods, while ChatIE and CodeIE show substantial degradation on both task types. GuideNER is applicable only to NER tasks.}
\label{fig:contrast}
\end{figure}

Guideline-based approaches offer an alternative to task transfer 
methods, but existing work has critical limitations. GuideNER~\citep{huang2025guidener}, the current state-of-the-art, has significant 
limitations. First, it uses simple frequency-based selection without systematic optimization for guideline quality. 
Second, it is designed specifically for NER and does not extend to other IE 
tasks, as evidenced by the absence of RE results in 
Figure~\ref{fig:contrast}. These limitations motivate the need for a more 
general and optimized approach.


Building on these observations, we introduce automatic subcategory generation 
(Figure~\ref{fig:motivation}, bottom) that decomposes labels into semantically 
discrete atomic representations. The key insight is that IE labels are often coarse-grained: a "Person" label in NER could mean family roles like "father" or "friend" to the model's prior understanding, while in a specific dataset it may only refer to public figures such as "athlete" or "politician". To bridge this gap between the model's prior knowledge and the dataset's annotation schema, we represent each label using multiple subcategories as atomic representations, with these subcategory patterns serving as rules to clarify the label's specific meaning in context. For example, in NER, "Person" can be represented by subcategories such as "athlete" and "public figure"; in RE, "[FRESNO, Located-In, Calif]" can be decomposed into "[city, spatial-containment, state]" and "[subregion, asymmetry, super-region]".
Crucially, by discretizing labels into semantic atomic units, we can treat them as controllable discrete variables. This enables us to optimize label representations using optimization algorithms such as particle filtering, where each rule is a particle with an associated weight, refined through iterative evaluation and Bayesian updates.


\begin{figure}[!htbp]
\centering
\includegraphics[width=\linewidth]{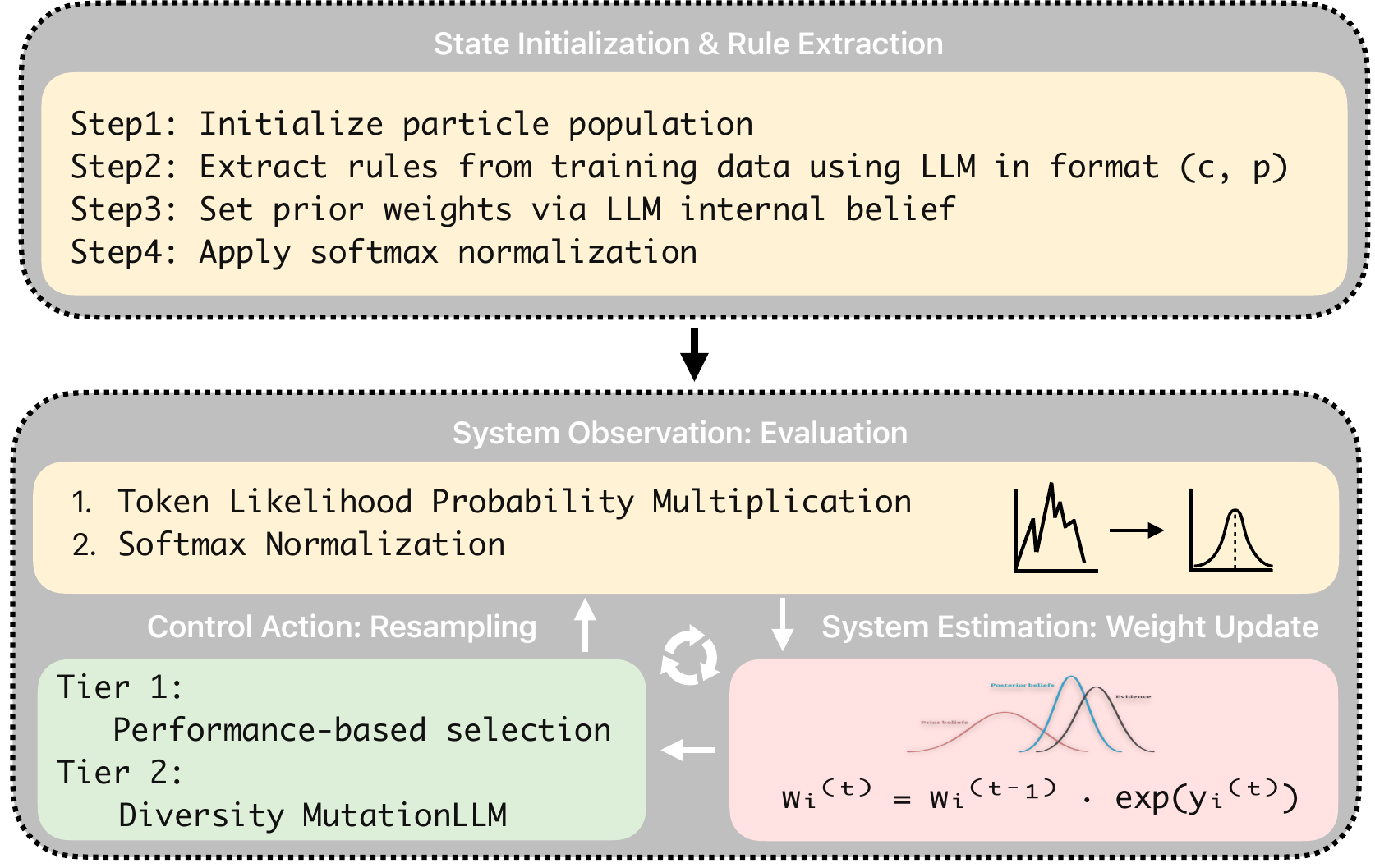}
\caption{Overview of the BCL framework. The framework operates through 
particle filtering with Bayesian updates, alternating between observation and control to progressively optimize the semantic patterns distribution.}
\label{fig:rulefilter_overview}
\end{figure}

We introduce \textbf{BCL (Bayesian In-Context Learning Framework for Information
Extraction)}, optimizing subcategory patterns through four steps (Figure~\ref{fig:rulefilter_overview}): (1) initialization---generate initial patterns and set prior weights, (2) observation---evaluate via ICL-based IE to compute likelihoods, (3) weight update---refine weights through Bayesian update, (4) resampling---eliminate low-weight particles and diversify high-performers via LLM mutation.

Our contributions are:
\begin{itemize}[itemsep=2pt]
    \item We introduce the key insight of treating context as controllable discrete 
variables, achieved by decomposing labels into fine-grained semantic units, 
enabling systematic optimization methods to be applied.
    
    \item We develop the first optimization framework using particle filtering with Bayesian updates that generalizes across IE tasks, achieving systematic quality improvement on both sequence labeling and relation classification paradigms.
    
    \item Extensive experiments demonstrate substantial improvements over existing approaches (up to 30\%), achieving strong performance while other methods either fail to generalize or show limited effectiveness.

    
    
\end{itemize}

\section{Related Work}
\subsection{In-Context Learning for Information Extraction}

In-context learning (ICL) enables large language models to adapt to new tasks through demonstration examples without parameter updates \citep{brown2020language,wei2022emergent,min2022rethinking}. Traditional ICL approaches for information extraction rely on example-based demonstrations, where models learn input-output mappings through pattern recognition \citep{dong2022survey,li2023unified}.

Recent work improves ICL for IE by refining demonstration construction, retrieval, and filtering strategies. C-ICL \citep{mo2024c} incorporates both positive and hard negative examples into demonstrations, while G\&O \citep{li2024simple} decomposes generation into intermediate reasoning and structured outputs to improve stability. GuideNER \citep{huang2025guidener} replaces demonstrations with LLM-generated annotation guidelines, and Dr.ICL \citep{WOS:001446914500003} retrieves task-relevant examples to enhance reasoning performance. Similarly, MAPS \citep{DOI:10.3724/2096-7004.di.2025.0025} introduces anchor-based sampling for fine-grained entity linking, while recent LLM-based feature selection methods \citep{DOI:10.3724/2096-7004.di.2025.0078} further highlight the importance of iterative filtering for structured extraction. Related observations also appear in adjacent multimodal understanding settings: Human Motion Instruction Tuning \citep{li2025human} and Multiple Human Motion Understanding \citep{li2026multiple} show that carefully designed instruction and structured semantic supervision can improve complex motion understanding, suggesting that input organization and guidance are broadly important for structured prediction.

Beyond demonstration design, recent studies analyze the intrinsic mechanisms of ICL and context utilization. \citet{shi2026intrinsic} study entropy in context length scaling, while \citet{cai2025role} examine the roles of deductive and inductive reasoning. \citet{lan2025attention} further propose attention consistency to estimate token importance, providing insights into how models utilize demonstrations during inference.

For relation extraction, ICL faces challenges in modeling inter-entity dependencies and contextual patterns. GPT-RE \citep{wan2023gpt} retrieves task-aware demonstrations with label-guided reasoning, while \citet{li2024recall} propose a recall–retrieve–reason framework to enhance retrieval and reasoning. \citet{wadhwa2023revisiting} highlight performance variance across prompts, and CodeIE \citep{li2023codeie} reformulates IE as code generation but remains sensitive to demonstration quality.


Recent studies show that LLMs can perform structured reasoning in complex settings. CountLLM \citep{yao2025countllm} highlights structured dependency modeling, while other work explores retrieval--reasoning in multi-hop QA \citep{ji2026retrieval}, few-shot generalization without explicit meta-learning \citep{guan2025meta,guan2025learning}, and structured context in visually grounded retrieval-augmented generation \citep{10.1145/3726302.3730070}. Together, these findings highlight the importance of context utilization.

\subsection{Control-Theoretic and Probabilistic Optimization}

Classical control theory \citep{aastrom2021feedback} models complex systems as input-output mappings governed by feedback mechanisms, where external control variables can systematically steer system behavior without directly observing internal states. Particle filtering \citep{gordon1993novel} and sequential Monte Carlo methods \citep{doucet2001sequential} estimate latent states in high-dimensional nonlinear systems via population-based sampling and importance resampling. In black-box optimization, Bayesian optimization \citep{frazier2018tutorial, Xu_Liu_Mattei_Zheng_2026} builds probabilistic surrogate models with acquisition functions to guide sampling, while Approximate Bayesian Computation \citep{beaumont2002approximate, liu2026discoveringcontrolinterventionalboundary} enables likelihood-free inference for complex models. These approaches share a common principle: optimizing system behavior via input-output observations without access to internal mechanisms. Evolutionary prompt optimization \citep{qi2024evolution} applies population-based search to LLM behavior, but lacks systematic control-theoretic grounding and focuses on reasoning tasks rather than structured prediction. In contrast, our work integrates control-theoretic principles with sequential Monte Carlo methods to optimize demonstration selection in few-shot learning.

Recent work begins to apply such principles to controlling LLM behavior. For example, \citet{cai2025bayesian} leverage Bayesian optimization to steer LLM-driven image editing processes under black-box settings, demonstrating the effectiveness of probabilistic search for controllable generation.

\subsection{Optimization Approaches for LLM Behavior}

Various optimization strategies have been explored for LLM behavior control\citep{Zhao_Min_Wu_Li_Sun_Cai_Wang_Chen_Penn_2026, cao2025pretraining}. Fine-tuning \citep{wei2021finetuned} requires substantial resources and labeled data, limiting few-shot applicability. Prompt engineering \citep{zhou2022large, pryzant2023automatic} relies on manual effort or local search heuristics, while evolutionary algorithms \citep{qi2024evolution} explore prompt spaces but lack systematic guideline optimization for structured prediction. Existing methods rely on heuristic strategies without control-theoretic grounding \citep{zhao2021calibrate}. Although RLHF \citep{ouyang2022training} and preference optimization \citep{rafailov2023direct} address alignment, they modify model parameters rather than optimizing external control inputs like demonstration selection rules.

\section{Methodology: BCL}
\label{sec:methodology}
\begin{figure*}[t]
\centering
\includegraphics[width=0.9\textwidth, trim=50 0 40 0, clip]{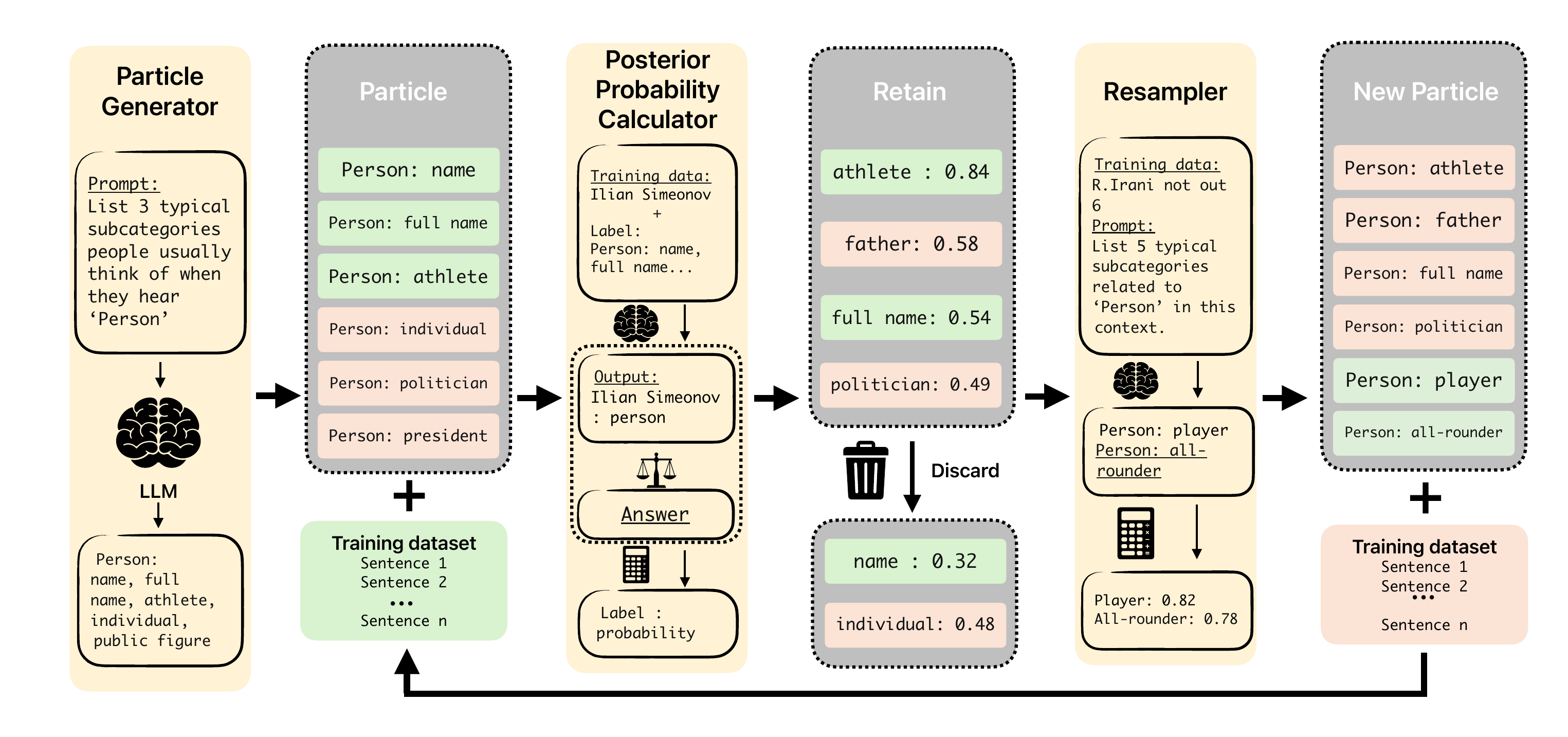}
\caption{Overall framework of BCL showing the particle-based rule optimization pipeline with iterative generation (Particle Generator), evaluation(Posterior Probability Calculator), selection (Retain), and mutation (Resampler) phases guided by LLM performance feedback.}
\label{fig:main}
\end{figure*}

Our BCL framework consists of a comprehensive control-theoretic approach for rule optimization, as illustrated in Figure~\ref{fig:rulefilter_overview} and Figure~\ref{fig:main}. Figure~\ref{fig:rulefilter_overview} shows the algorithmic overview with the four key steps of our adaptive filtering process, while Figure~\ref{fig:main} presents the particle-based optimization pipeline with iterative generation, evaluation, selection, and mutation phases.

\subsection{Problem Formulation}

Given a pre-trained large language model $\mathcal{M}$ and a target dataset $\mathcal{D} = \{(x_i, y_i)\}_{i=1}^N$ with training split $\mathcal{D}_{train}$, development split $\mathcal{D}_{dev}$ and test split $\mathcal{D}_{test}$, our goal is to find an optimal rule list $\mathcal{R}^*$ that maximizes the model's information extraction performance on the development set, and then evaluate its generalization performance on the test set.

\subsubsection{Dataset-Specific Optimization}
Since different datasets have varying data distributions and annotation conventions, we need to adapt the rule selection to each specific dataset. Using the training portion $\mathcal{D}_{train}$, we extract candidate rules, and then optimize the rule list $\mathcal{R}^*$ based on performance on the development set $\mathcal{D}_{dev}$. This ensures that the final rule list is tailored to both the dataset characteristics and the specific LLM's behavior.

\subsubsection{ICL-based Extraction}
For any input $x$, the extraction process follows:
\begin{align}
\hat{y} = \mathcal{M}(\text{Prompt}(x, R_t))
\end{align}
where $\mathcal{M}$ is the pre-trained LLM, $\text{Prompt}(x, R_t)$ constructs the input prompt by combining text $x$ with the optimal rule configuration $R_t$, and $\hat{y}$ is the predicted extraction output.

\subsubsection{Optimization Objective}
We seek to find the optimal rule configuration that maximizes performance on unseen data. Following standard machine learning practice, we split the available training data into training and validation sets, and optimize rules based on validation performance:

\begin{align}
R_t^* &= \argmax_{R_t} \frac{1}{|\mathcal{D}_{\text{val}}|} \sum_{(x,y) \in \mathcal{D}_{\text{val}}} F(\hat{y}, y)
\end{align}

where $R_t = \{(p_i^{(t)}, c_i, w_i^{(t)})\}_{i=1}^N$ represents a rule configuration with $N$ particles at iteration $t$, where $p_i^{(t)}$ is the $i$-th subcategory pattern (particle), $c_i$ is its corresponding entity label, and $w_i^{(t)}$ is the confidence weight, $F(\cdot, \cdot)$ is a performance metric (e.g., F1 score).
The rule extraction and initial population are derived from $\mathcal{D}_{\text{train}}$, while the optimization objective is evaluated on the held-out validation set $\mathcal{D}_{\text{val}}$ to prevent overfitting. Final evaluation is performed on $\mathcal{D}_{\text{test}}$.

\subsubsection{Challenges with Existing Approaches}
While existing rule-based ICL methods like GuideNER have demonstrated the effectiveness of annotation guidelines over examples, they lack systematic optimization strategies for rule selection and combination. Current approaches rely on heuristic frequency-based filtering, which fails to capture the complex interdependencies between rules and their collective impact on IE performance.

\subsubsection{Control-Theoretic Reformulation}
Traditional LLM optimization faces a fundamental challenge: the massive parameter space (billions of parameters) renders direct observation and control intractable. We address this by treating rules as a low-dimensional, observable interface to the LLM system.

Recent findings suggest that in-context learning operates as a rule-based inference system, where the quality of rules—not examples—determines performance. This motivates reformulating the optimization as a control system problem, where rules serve as externally controllable state variables:

\begin{align}
R_{t+1} &= f(R_t, u_t) \quad \text{(Rule Evolution)} \\
y_t &= h(R_t) \quad \text{(Performance Observation)}
\end{align}
where $R_t$ represents the rule configuration at iteration $t$, $u_t$ denotes control actions (rule modifications), and $y_t$ is the observed IE performance.

\subsection{Adaptive Rule Filtering Algorithm}

The discrete, combinatorial nature of rule spaces and nonlinear performance mappings makes classical control methods inappropriate. We develop an adaptive filtering approach that iteratively estimates optimal rule configurations through performance feedback.

As illustrated in Figure~\ref{fig:rulefilter_overview}, our approach follows a systematic particle filtering process:

\subsubsection{Particle-based State Representation}

\textbf{Terminology.} To clarify key concepts used throughout this section:
\begin{itemize}
    \item \textbf{Particle}: A subcategory pattern paired with its entity label, 
    e.g., (``athlete'', Person)
    \item \textbf{Rule}: A particle with its associated confidence weight in the population
    \item \textbf{Weight}: Normalized probability $w_i \in [0,1]$ indicating particle quality, 
    where $\sum_i w_i = 1$
\end{itemize}

At time step $t$, the complete rule configuration is:
\begin{align}
R_t = \{(p_i^{(t)}, c_i, w_i^{(t)})\}_{i=1}^{N} \label{eq:state_representation}
\end{align}
where $p_i^{(t)}$ is the $i$-th subcategory pattern (particle), $c_i$ is its corresponding entity label, and $w_i^{(t)}$ is the confidence weight.


The filtering process consists of four steps:

\textbf{0. Initialization (Rule Extraction):}
We initialize the particle population by extracting initial rules from the 
training dataset using the GuideNER approach~\citep{huang2025guidener}. For 
each input-label pair $(x_j, y_j)$ in the training set, we use the LLM to 
summarize rule patterns:
\begin{align}
p_i^{(0)} = \text{LLM}_{\text{extract}}(x_j, y_j, \text{prompt}_{\text{summary}}) \label{eq:rule_extraction}
\end{align}

The initial particles are assigned prior weights based on their linguistic 
naturalness under the LLM's language model. We generate $N$ particles per entity 
label (typically $N=10$) and compute their prior scores:
\begin{align}
s_i &= -\text{PPL}(p_i^{(0)}), \quad i=1,\ldots,N \label{eq:init_scores}\\
w_i^{(0)} &= \frac{\exp(s_i)}{\sum_{j=1}^{N} \exp(s_j)} \label{eq:init_weights}
\end{align}
where perplexity is computed as:
\begin{equation}
\text{PPL}(p_i^{(0)}) = \exp\left(-\frac{1}{|p_i^{(0)}|} \sum_{j=1}^{|p_i^{(0)}|} 
\log P(t_j | t_{<j})\right) 
\label{eq:perplexity}
\end{equation}
Lower perplexity indicates the rule text is more natural and coherent according 
to the model's internal knowledge, thus receiving higher prior weight.

\textbf{1. Evaluation (ICL-based Observation):}
Each particle is evaluated through ICL-based IE inference on a validation batch 
$\mathcal{B}^{(t)}$ sampled from $\mathcal{D}_{\text{train}}$. For each particle, 
we compute their confidence scores:
\begin{align}
s_i^{(t)} &= \text{L}_{\text{c}}(p_i^{(t)}, \theta), \quad i=1,\ldots,N \label{eq:eval_scores}\\
y_i^{(t)} &= \frac{\exp(s_i^{(t)})}{\sum_{j=1}^{N} \exp(s_j^{(t)})} \label{eq:eval_weights}
\end{align}
where $\text{L}_{\text{c}}(p_i^{(t)}, \theta)$ computes the average 
log probability when the LLM uses rule $p_i^{(t)}$ to generate a label sequence 
$T_i$ on a validation sample:
\begin{equation}
\text{L}_{\text{c}}(p_i, \theta) = \frac{1}{|T_i|} \sum_{j=1}^{|T_i|} \log P(t_j | t_{<j}, p_i, \theta) 
\label{eq:confidence_definition}
\end{equation}
Here $\theta$ represents the pretrained LLM parameters. This provides a 
length-normalized confidence score $y_i^{(t)} \in (0,1]$ reflecting how well 
the rule performs on the actual IE task.

\textbf{2. Weight Update (Bayesian Posterior):}
Particle weights are updated by combining prior knowledge with observed performance:
\begin{align}
\tilde{w}_i^{(t)} &= w_i^{(t-1)} \cdot \exp(\beta \cdot y_i^{(t)}) 
\label{eq:update_unnorm}\\
w_i^{(t)} &= \frac{\tilde{w}_i^{(t)}}{\sum_{j=1}^{N} \tilde{w}_j^{(t)}} 
\label{eq:update_norm}
\end{align}
where $w_i^{(t-1)}$ is the prior weight and $\exp(\beta \cdot y_i^{(t)})$ rewards 
higher performance.

\textbf{3. Multi-level Resampling with Rule Mutation:}
We employ a two-tier strategy balancing exploitation and exploration, adapting 
the approach from \citet{qi2024evolution}:

\textbf{Tier 1 (Performance Selection):} Retain the top 50\% of particles by weight, 
removing low-performing ones.

\textbf{Tier 2 (Diversity Mutation):} Apply semantic mutations to retained 
particles through LLM-guided generation:
\begin{equation}
p_i' = \text{LLM}_{\text{mutate}}(p_i^{(t-1)}, x^{(t)}) 
\label{eq:mutation}
\end{equation}
where $x^{(t)}$ provides context. Three strategies are employed: 
\textit{refinement} (increase specificity), \textit{generalization} 
(increase coverage), and \textit{contextualization} (generate domain-specific variants).

New particles inherit labels from parents and receive weights based on perplexity 
as described in Equation~\ref{eq:perplexity}:
\begin{align}
s_i' &= -\text{PPL}(p_i'), \quad i=1,\ldots,M 
\label{eq:resample_scores}\\
w_i' &= \frac{\exp(s_i')}{\sum_{j=1}^{M} \exp(s_j')} 
\label{eq:resample_weights}
\end{align}
Lower perplexity indicates greater consistency with the model's language patterns (See Appendix~\ref{app:weight} for the rationale behind our prior and likelihood selection in the Bayesian filtering framework.).
The updated configuration combines retained and new particles: 
$R^{(t)} = \mathcal{R}_{\text{keep}} \cup \mathcal{R}_{\text{new}}$.

The iteration continues until: (a) performance saturates with 
relative improvement $(F_1^{(t)} - F_1^{(t-3)})/F_1^{(t-3)} <  0.03$ 
over 3 iterations, or (b) all training samples are exhausted. 

As illustrated in Table~\ref{tab:training_size}, convergence typically occurs 
with only 3--5\% of training data when performance plateaus.
\section{Experiment}

We conduct extensive experiments on multiple Information Extraction tasks. Following previous work~\citep{huang2025guidener, wei2023chatie, li2023codeie}, we use entity-level F1 score for evaluation. While our framework is applicable to various IE paradigms, we focus on two representative tasks: sequence labeling (NER) and relation classification (RE). For NER, this requires correct boundary detection and type classification; for RE, correct identification of both entity arguments and their relation type (see Appendix~\ref{app:evaluation} for precise definitions). Following GuideNER~\citep{huang2025guidener}, we measure token cost as the average number of input and output tokens per sample, reflecting computational overhead and inference latency.


\subsection{Datasets}

We evaluate on six widely-used IE benchmarks spanning multiple domains and task formulations. For NER, we use CoNLL-2003~\citep{tjong-kim-sang-de-meulder-2003-introduction} (news, 4 entity types), ACE 2005~\citep{walker2006ace} (news/conversational, 7 types), and GENIA~\citep{kim2003genia} (biomedical, 5 types). For RE, we use NYT~\citep{riedel2010modeling} (news, 24 relation types), CoNLL04~\citep{roth2004linear} (general domain, 5 types), and SciERC~\citep{zhang2024scier} (scientific papers, 7 types). Appendix~\ref{app:dataset_stats} provides detailed statistics.

\begin{table*}[htbp]
\centering
\footnotesize
\setlength{\tabcolsep}{9pt}
\begin{tabular}{llccccccr}
\toprule
\multirow{2}{*}{Method} & \multirow{2}{*}{Model} 
& \multicolumn{3}{c}{NER} 
& \multicolumn{3}{c}{RE} 
& \multirow{2}{*}{\begin{tabular}[c]{@{}c@{}}Token\\Cost\end{tabular}} \\
\cmidrule(lr){3-5} \cmidrule(lr){6-8}
& & CoNLL03 & ACE05 & GENIA & NYT & CoNLL04 & SciERC & \\
\midrule
\multirow{4}{*}{One-shot} 
& Qwen-2.5-3b & 60.55 & 27.08 & 46.78 & 0.29 & 19.28 & 5.01 & \multirow{4}{*}{385} \\
& Qwen-2.5-7b & 62.82 & 35.73 & 52.91 & 0.28 & 28.10 & 5.54 & \\
& Llama-3.1-8b & 65.78 & 40.98 & 51.29 & 0.44 & 22.57 & 7.18 & \\
& Pixtral-12B & 60.25 & 39.15 & 49.73 & 0.38 & 21.70 & 7.86 & \\
\midrule

\multirow{4}{*}{ChatIE} 
& Qwen-2.5-3b & 39.26 & 15.41 & 33.32 & 0.00 & 11.11 & 0.00 & \multirow{4}{*}{942} \\
& Qwen-2.5-7b & 25.64 & 11.83 & 10.19 & 0.00 & 12.48 & 0.88 & \\
& Llama-3.1-8b & 55.12 & 26.91 & 35.20 & 0.00 & 12.62 & 1.03 & \\
& Pixtral-12B & 60.85 & 29.28 & 40.12 & 0.40 & 14.50 & 4.40 & \\
\midrule

\multirow{4}{*}{CodeIE} 
& Qwen-2.5-3b   & 45.92 & 23.70  & 6.59  & 0.00 & 0.00 & 0.00 & \multirow{4}{*}{1172} \\
& Qwen-2.5-7b   & 60.00 & 22.76  & 19.17  & 0.00 & 0.00 & 0.00 & \\
& Llama-3.1-8b  & 0.05 & 0.06 & 0.00 & 0.00 & 0.00 & 0.00 & \\
& Pixtral-12B   & 53.39 & 18.57  & 20.62  & 0.00 & 0.00 & 0.00 & \\
\midrule

\multirow{4}{*}{GuideNER} 
& Qwen-2.5-3b & 63.32 & 27.43 & 41.49 & — & — & — & \multirow{4}{*}{506} \\
& Qwen-2.5-7b & 65.10 & 41.57 & 47.43 & — & — & — & \\
& Llama-3.1-8b & 61.38 & 44.97 & 42.86 & — & — & — & \\
& Pixtral-12B & 64.76 & 37.64 & 48.03 & — & — & — & \\
\midrule

\multirow{4}{*}{BCL} 
& \cellcolor{gray!20}Qwen-2.5-3b 
& \cellcolor{gray!20}65.12 & \cellcolor{gray!20}35.46 & \cellcolor{gray!20}46.98 
& \cellcolor{gray!20}0.31 & \cellcolor{gray!20}35.32 & \cellcolor{gray!20}8.28 
& \multirow{4}{*}{501} \\
& \cellcolor{gray!20}Qwen-2.5-7b 
& \cellcolor{gray!20}72.83 & \cellcolor{gray!20}46.94 & \cellcolor{gray!20}51.36 
& \cellcolor{gray!20}0.43 & \cellcolor{gray!20}42.46 & \cellcolor{gray!20}9.57 & \\
& \cellcolor{gray!20}Llama-3.1-8b 
& \cellcolor{gray!20}69.14 & \cellcolor{gray!20}53.10 & \cellcolor{gray!20}50.15 
& \cellcolor{gray!20}0.91 & \cellcolor{gray!20}38.93 & \cellcolor{gray!20}12.06 & \\
& \cellcolor{gray!20}Pixtral-12B 
& \cellcolor{gray!20}65.54 & \cellcolor{gray!20}42.87 & \cellcolor{gray!20}50.65 
& \cellcolor{gray!20}0.60 & \cellcolor{gray!20}25.75 & \cellcolor{gray!20}10.34 & \\
\bottomrule
\end{tabular}

\caption{Performance comparison of different methods (One-shot, ChatIE, CodeIE, GuideNER, and BCL) across Qwen-2.5, Llama-3.1, and Pixtral models on NER (CoNLL03, ACE05, GENIA) and RE (NYT, CoNLL04, SciERC) benchmarks. All results are statistically significant ($p < 0.05$). Gray shading highlights the best-performing method for each model configuration.}

\label{tab:method_performance}
\end{table*}

\subsection{Experiments Setup}
All experiments are conducted on a computing cluster equipped with H100 GPUs for computational acceleration.
They are implemented using PyTorch 2.6.0 and transformers 4.51.3. We employ four foundation models with varying scales, languages, and training datasets: Qwen2.5-3B, Qwen2.5-7B, Llama3.1-8B and Pixtral-12B. This selection allows us to investigate the impact of different model size on our method's performance. All models are tested with temperature set to 0.0 and random seed fixed at 42 to ensure reproducibility of the experiments. 
In addition, during the computation of prior probabilities, all models are evaluated in eval mode to disable dropout and other stochastic behaviors.

For optimization framework, we set the number of particles to 10 based on empirical experience from preliminary experiments(~\citep{huang2025guidener}), balancing computational efficiency and exploration capability. The number of data used in each observation step is determined through grid search over the set [1, 3, 5, 7, 9, 11, 13, 15], selecting the value that yields optimal performance for each dataset.
This configuration ensures consistent experimental conditions and enables direct performance comparison across different models and datasets.

\subsection{Baselines}

We compare against widely-used ICL approaches in the IE domain. 

Our baselines include: \textbf{One-shot}, where models perform IE with a single demonstration example to illustrate the task format and desired output structure.\textbf{ChatIE}~\citep{wei2023chatie} 
and \textbf{CodeIE}~\citep{li2023codeie}, two task transfer methods that 
reformulate IE as dialogue or code generation tasks, respectively. Both methods 
employ sophisticated example selection strategies and are applicable to both NER 
and RE tasks. \textbf{GuideNER}~\citep{huang2025guidener}, the current 
state-of-the-art guideline-based method for NER, which uses frequency-based rule 
selection to assist in-context learning. Since GuideNER is specifically designed 
for NER tasks, we only evaluate it on NER datasets and mark it as "—" for RE tasks.

This experimental design ensures all baseline methods operate under the same 
paradigm of pure in-context learning, enabling fair comparison of different 
approaches' effectiveness.

\subsection{Main Results}
Table~\ref{tab:method_performance} shows BCL consistently outperforms 
all baselines across IE benchmarks and model scales. Task transfer 
baselines (ChatIE, CodeIE) exhibit severe degradation on smaller models: 
ChatIE achieves only 25.64 F1 on CoNLL03 (Qwen-2.5-7B) versus BCL's 
72.83, while CodeIE drops to 0.05 F1 (Llama-3.1-8B) versus BCL's 69.14, 
reflecting their reliance on model-specific capabilities. Against the 
rule-based GuideNER, BCL shows consistent advantages (72.83 vs. 65.10 
on CoNLL03; 51.36 vs. 47.43 on GENIA), demonstrating systematic Bayesian 
optimization's superiority over frequency heuristics.

BCL's advantage amplifies on RE tasks where prior methods fail completely. 
ChatIE and CodeIE achieve 0.00 F1 across most RE configurations, while 
GuideNER is inapplicable (marked "—"). In contrast, BCL maintains effective 
performance (42.46 F1 on CoNLL04; 12.06 F1 on SciERC), demonstrating 
successful generalization across both NER and RE through adaptive rule 
optimization.

This stability enables cost-efficient deployment: Qwen-2.5-3B with BCL 
(65.12 F1) matches Llama-3.1-8B's one-shot performance (65.78 F1) on 
CoNLL03, achieving comparable results with 62\% fewer parameters. BCL thus represents the first ICL optimization framework leveraging Bayesian inference for adaptive optimization, achieving consistent performance across diverse IE paradigms and model scales while overcoming the task-specific limitations of prior methods.(Appendix~\ref{appendix:case_study} illustrates the evolution process 
through examples).

\subsection{Ablation Study}

We conduct ablation studies to validate the contribution of each component in BCL's particle filtering framework. Table~\ref{tab:ablation} presents results on CoNLL03 with Qwen-2.5-7B, where we systematically remove each mechanism while keeping others intact.
\begin{table}[t]
\centering
\caption{Ablation study on different components of the proposed method.}
\label{tab:ablation}
\begin{tabular}{p{0.42\columnwidth}cc}
\toprule
\textbf{Variant} & \textbf{F1 Score} & \textbf{Particles} \\
\midrule
BCL (Full) & 72.83 & 10 \\
Weight Update & 61.78 & 10 \\
Tier-1 Resampling & 71.29 & 10,441 \\
Tier-2 Resampling & 62.98 & 10 \\
\bottomrule
\end{tabular}
\end{table}

\textbf{Bayesian Weight Update.} Removing Bayesian weight updates causes the most significant performance drop (-11.05 F1 points), reducing F1 from 72.83 to 61.78. Without this mechanism, the framework degenerates to random search without principled belief updates, demonstrating that Bayesian inference is the core component enabling BCL's effectiveness.

\textbf{Tier-2 Resampling (Diversity Mutation).} Disabling diversity mutation leads to the second-largest degradation (-9.85 F1 points), with F1 dropping to 62.98. This confirms that maintaining particle diversity through controlled mutation is crucial for preventing premature convergence to suboptimal solutions and exploring the prompt space effectively.

\textbf{Tier-1 Resampling (Performance Filtering).} While removing this component causes minimal performance loss (-1.54 F1 points), particle count explodes from 10 to 10,441, revealing that Tier-1 primarily ensures efficiency by pruning low-quality particles with negligible performance cost.

These results validate BCL's design: Bayesian weight updates and diversity mutation are essential for performance, while performance filtering ensures efficiency by maintaining a compact particle set without performance degradation.

\textbf{Effect of Semantic Decomposition.} 
To isolate the role of semantically coherent subcategories, we conduct an additional ablation where subcategories are randomly reassigned to different entity labels, breaking semantic alignment while keeping the subcategory pool unchanged. 
Results (Appendix~\ref{app:semantic_ablation}) show that this leads to consistent performance degradation, confirming that semantic coherence is critical for effective optimization.

\subsection{Cross-Model Generalization}

While our method is designed to optimize inference on a given model, 
it is also important to understand whether the learned rules capture transferable patterns 
that extend beyond a specific backbone. 
To this end, we study the cross-model generalization ability of the proposed approach.

We consider a setting where rules optimized on smaller open-source models 
(e.g., Llama-3.1-8B and Qwen-2.5-7B) are directly applied to a stronger closed-source model, 
GPT-3.5-turbo. This setup reflects a practical scenario in which optimization is performed 
on accessible models and then deployed on more capable systems. 
The inference pipeline remains unchanged, and no additional adaptation is introduced.

\begin{table}[h]
\centering
\small
\setlength{\tabcolsep}{4pt}
\begin{tabular}{l l c c}
\toprule
\textbf{Summary Model} & \textbf{Inference Model} & \textbf{CoNLL03} & \textbf{ACE05} \\
\midrule
-- & GPT-3.5-turbo & 73.44 & 51.03 \\
\midrule
Llama-3.1-8B & Llama-3.1-8B & \textbf{69.14} & \textbf{53.10} \\
Llama-3.1-8B & GPT-3.5-turbo & \textbf{74.60} & \textbf{52.50} \\
\midrule
Qwen-2.5-7B & Qwen-2.5-7B & \textbf{72.83} & \textbf{46.94} \\
Qwen-2.5-7B & GPT-3.5-turbo & \textbf{75.80} & \textbf{53.00} \\
\bottomrule
\end{tabular}
\caption{Cross-model generalization results. Rules optimized on smaller open-source models 
can be directly transferred to GPT-3.5-turbo and consistently improve performance.}
\label{tab:cross_model}
\end{table}

As shown in Table~\ref{tab:cross_model}, rules learned on smaller models 
consistently yield performance gains when transferred to GPT-3.5-turbo. 
In particular, the transferred rules outperform the direct GPT-3.5-turbo baseline 
on both datasets. This suggests that the proposed method captures 
generalizable reasoning patterns rather than relying on model-specific behaviors.

Overall, these results indicate that the learned optimization strategies 
are not limited to the source model, but can extend to stronger models 
without additional tuning.

\subsection{Parameter Sensitivity Analysis}
We perform extensive parameter analysis on two key factors: context window length and training data size.

\subsubsection{Impact of Training Data Quantity on BCL Performance}
\label{main:quantity}
\begin{table}[h]
\centering
\caption{F1 scores (\%) for different training set sizes.}
\label{tab:training_size}
\begin{tabular}{>{\centering\arraybackslash}p{1.8cm}ccc}
\toprule
\textbf{Training Set Size} & \textbf{CoNLL03} & \textbf{GENIA} & \textbf{ACE05} \\
\midrule
1\%  & 69.12 & 44.32 & 40.32 \\
3\%  & 71.16 & 50.12 & 44.04 \\
5\%  & 70.78 & 51.20 & 46.94 \\
10\% & 71.12 & 51.36 & 44.43 \\
20\% & 72.04 & 49.58 & 44.89 \\
30\% & 71.37 & 50.18 & 44.98 \\
40\% & 72.83 & 49.18 & 44.62 \\
50\% & 71.81 & 49.35 & 44.27 \\
\bottomrule
\end{tabular}
\end{table}


We investigate BCL's data efficiency by evaluating performance across varying training set sizes (1\% to 50\%). Table 3 presents F1 scores on Qwen-2.5-7B across three datasets with varying training set sizes. The results reveal a striking pattern: performance rapidly saturates with minimal data. On CoNLL03, BCL reaches 69.12 F1 with only 1\% of training data and peaks at 72.83 F1 with 40\%, showing modest improvement (~3.7 points) when scaling from 1\% to the optimal setting. Similar trends appear on GENIA (peak at 5-10\%: 51.20-51.36 F1) and ACE05 (peak at 5\%: 46.94 F1). This data efficiency stems from the particle filtering mechanism's ability to rapidly update belief distributions after each sample, achieving convergence without overfitting to large training sets.


\subsubsection{Impact of Context Window Length on BCL Performance}
\label{app:context_length}
\begin{table}[h]
\centering
\caption{F1 scores (\%) for different context lengths (in sentences) during the observation step}
\label{tab:context_length}
\begin{tabular}{>{\centering\arraybackslash}p{1.8cm}ccc}
\toprule
\textbf{Context Length} & \textbf{CoNLL03} & \textbf{GENIA} & \textbf{ACE05} \\
\midrule
1  & 63.37 & 45.20 & 38.15 \\
2  & 68.25 & 48.60 & 42.30 \\
4  & 72.04 & 51.80 & 46.95 \\
6  & 71.10 & 50.90 & 46.50 \\
8  & 71.04 & 50.20 & 46.20 \\
10 & 60.20 & 47.30 & 39.80 \\
12 & 59.85 & 46.80 & 39.20 \\
14 & 60.15 & 47.10 & 39.60 \\
\bottomrule
\end{tabular}
\end{table}

The context window length (also called batch size) determines how much information our method considers when computing posterior probabilities at each observation step. Table \ref{tab:context_length} shows that on Qwen-2.5-7B, performance improves significantly from single-sentence to five-sentence contexts due to smoother joint likelihood functions that provide more stable gradients. However, performance declines beyond 9 sentences as excessive observations lead to overly averaged particle weights, preventing effective updates.

\section{Conclusion}
In this paper, we propose BCL, a general framework that can efficiently traverse training sets and rapidly extract specific relationships from training data using particle-based methods. BCL is the first optimization framework using Bayesian inference for IE tasks. We have conducted extensive and comprehensive experiments to thoroughly demonstrate the effectiveness of our approach. 
Additionally, we acknowledge the limitations of our method, such as the generation of a large number of particles to achieve rapid convergence, which increases inference burden and computational costs in practice. Therefore, our future research will focus on improving particle utilization efficiency to further enhance the overall system performance.

\section*{Acknowledgement}
This work was supported in part by the National Key R\&D Program under grants 2025ZD1502903 and 2024YFC3308101.

\section*{Limitations}

Our work has several limitations. First, while BCL maintains reasonable 
inference efficiency, the optimization phase requires $O(K \times M)$ 
iterations, where $K$ denotes the number of particles and $M$ denotes the 
number of training samples. Each iteration involves CPU-intensive numerical 
computations including weight updates and resampling (detailed analysis in 
Appendix~\ref{app:complexity}). This makes the method most suitable for 
scenarios where optimization costs can be amortized across multiple deployments, 
rather than one-time or frequently-updated applications.

Second, the particle filtering mechanism may bias optimization toward frequent 
patterns in the training data, potentially affecting rare entity types. 
However, our empirical analysis (Appendix~\ref{app:long_tail}) shows that 
BCL is largely robust to such frequency imbalance, with only marginal 
performance differences across frequency strata.



\bibliography{latex/main}

\appendix

\section{Dataset Statistics}
\label{app:dataset_stats}

We provide detailed statistics for all six datasets used in our experiments in 
Table~\ref{tab:dataset_stats}.

\begin{table}[h]
\centering
\small  
\caption{Statistics of the datasets used in our experiments. $|$Ents$|$ and 
$|$Rels$|$ denote the number of entity types and relation types. \#Train, 
\#Val and \#Test denote the sample number in each split.}
\label{tab:dataset_stats}
\begin{tabular}{l|cc|rrr}
\toprule
\textbf{Dataset} & \textbf{$|$Ents$|$} & \textbf{$|$Rels$|$} & \textbf{\#Train} & \textbf{\#Val} & \textbf{\#Test} \\
\midrule
\multicolumn{6}{c}{\textit{Named Entity Recognition}} \\
\midrule
CoNLL03  & 4 & - & 14,041 & 3,250 & 3,453 \\
ACE05    & 7 & - & 6,202  & 745   & 812 \\
GENIA    & 5 & - & 15,023 & 1,669 & 1,854 \\
\midrule
\multicolumn{6}{c}{\textit{Relation Extraction}} \\
\midrule
NYT      & - & 24 & 56,195 & 5,000 & 5,000 \\
CoNLL04  & 4 & 5  & 922    & 231   & 288 \\
SciERC   & 6 & 7  & 1,861  & 275   & 551 \\
\bottomrule
\end{tabular}
\end{table}

\section{Case Study}

\label{appendix:case_study}
We present a case study using a CoNLL-2003 sentence to illustrate our three-stage workflow: particle generator, posterior probability calculator, and resampler. Here, $x$ represents label sequences, $N$ denotes the number of generation rules, and $Y$ represents input text. It is worth noting that the following case studies present simplified examples to illustrate our framework's workflow. All experiments follow the technical specifications detailed in Section~\ref{sec:methodology}.

\subsection{Case Study 1: Particle Generator}
\begin{figure*}[h]
\centering
\includegraphics[width=0.9\textwidth, trim=50 0 40 0, clip]{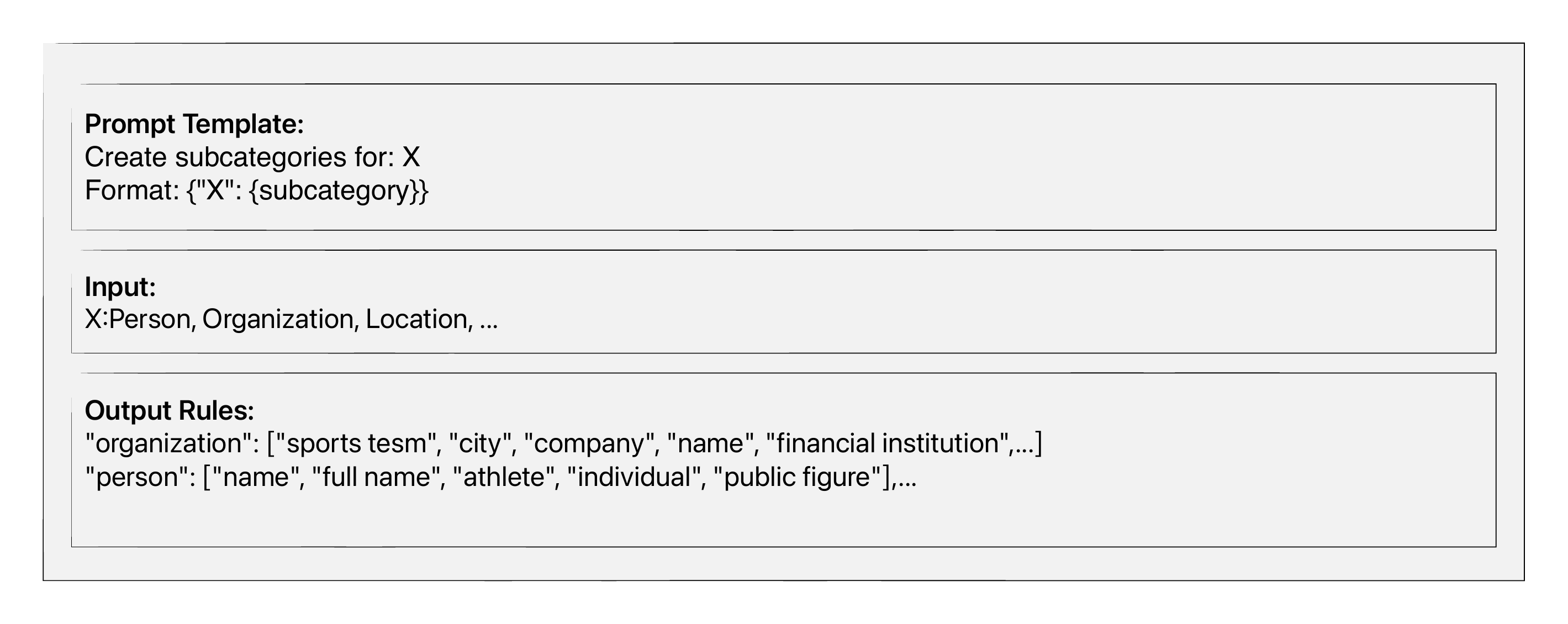}
\caption{Case study showing the prompt template for subcategory generation. The template guides the model to generate semantic subcategories for named entity labels, demonstrated with organization and person entity types.}
\label{fig:case1}
\end{figure*}
Particle generation serves as the initial step in our framework to obtain initial particles and compute their corresponding prior probabilities. We demonstrate this process using the label "organization" as an example. As shown in Figure\ref{fig:case1}, our prompt template instructs the model to act as a subcategory generation expert, decomposing broad entity labels into semantically distinct subcategories with associated probability weights.
For the "organization" label, the system generates diverse subcategories such as "sports team", "city", "company", "name", and "financial institution". Each subcategory represents a different semantic interpretation of how organizations might appear in text, with the probability weights serving as prior probabilities that reflect their typical occurrence frequency.

\subsection{Case Study 2: Posterior Probability Calculator}


\begin{figure*}[h]
\centering
\includegraphics[width=0.9\textwidth, trim=50 0 40 0, clip]{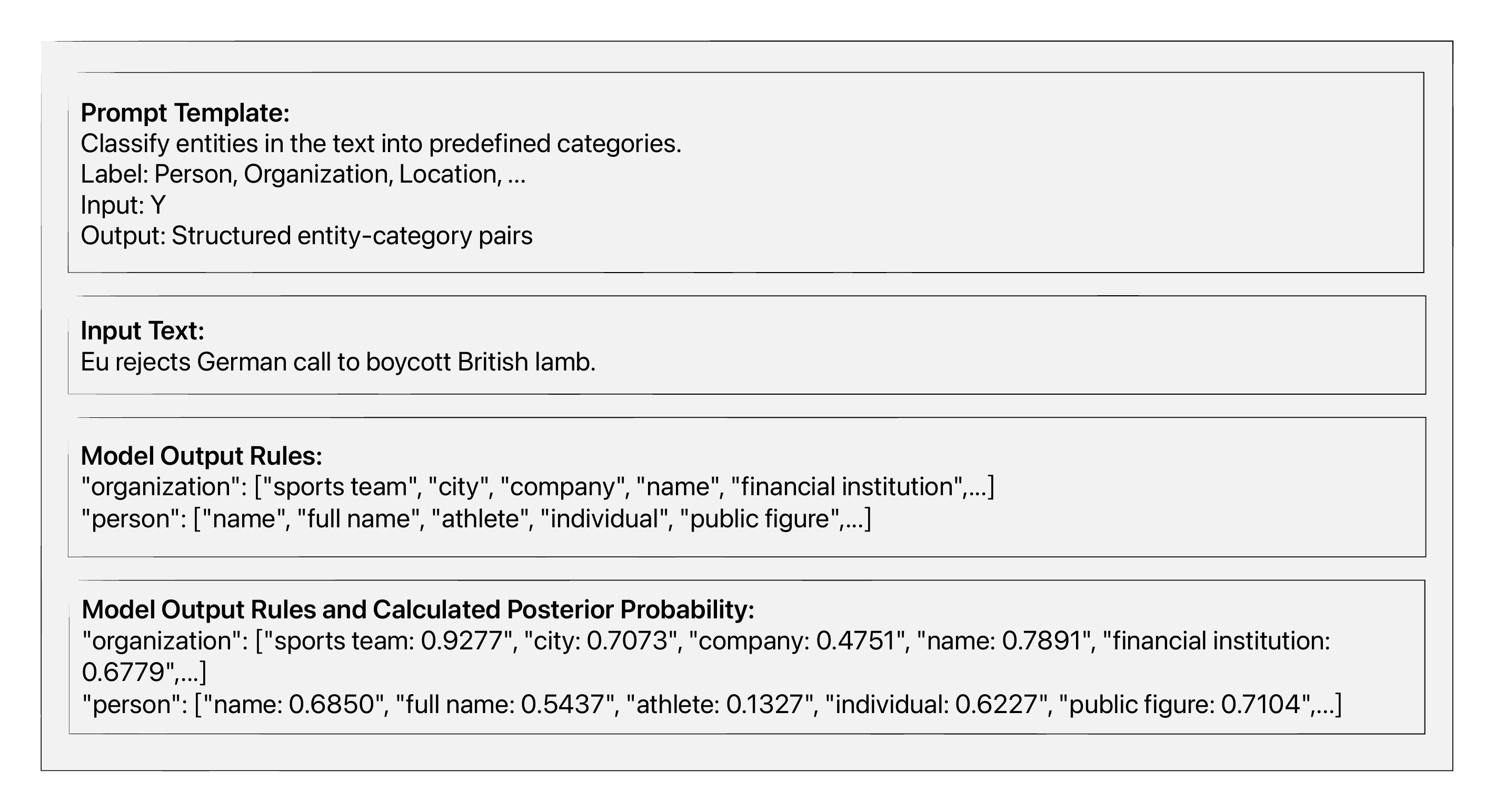}
\caption{Case study demonstrating posterior probability calculation through likelihood estimation. The model evaluates generated rules against input text to compute particle posterior probabilities based on classification accuracy.}
\label{fig:case2}
\end{figure*}
Following particle generation, we compute posterior probabilities by evaluating how well each generated rule performs on the actual classification task. Figure\ref{fig:case2} illustrates this process using the input text "EU rejects German call to boycott British lamb."
The system applies the previously generated subcategory rules to identify entities in the given text. The likelihood of each particle is calculated based on the model's ability to correctly identify and classify entities according to the generated rules.
The posterior probability calculation applies Bayes' theorem to update particle weights based on how well each rule matches the observed entity patterns in the input text. For instance, the "organization" subcategories show varying performance scores: "sports team: 0.9277", "city: 0.7073", "company: 0.4751", demonstrating how different semantic interpretations receive different posterior weights based on their effectiveness.
This likelihood-based evaluation ensures that particles with better classification performance receive higher posterior probabilities, enabling the system to focus on the most promising labeling hypotheses for subsequent resampling.

\subsection{Case Study 3: Resampler}

\begin{figure*}[h]
\centering
\includegraphics[width=0.9\textwidth, trim=50 0 40 0, clip]{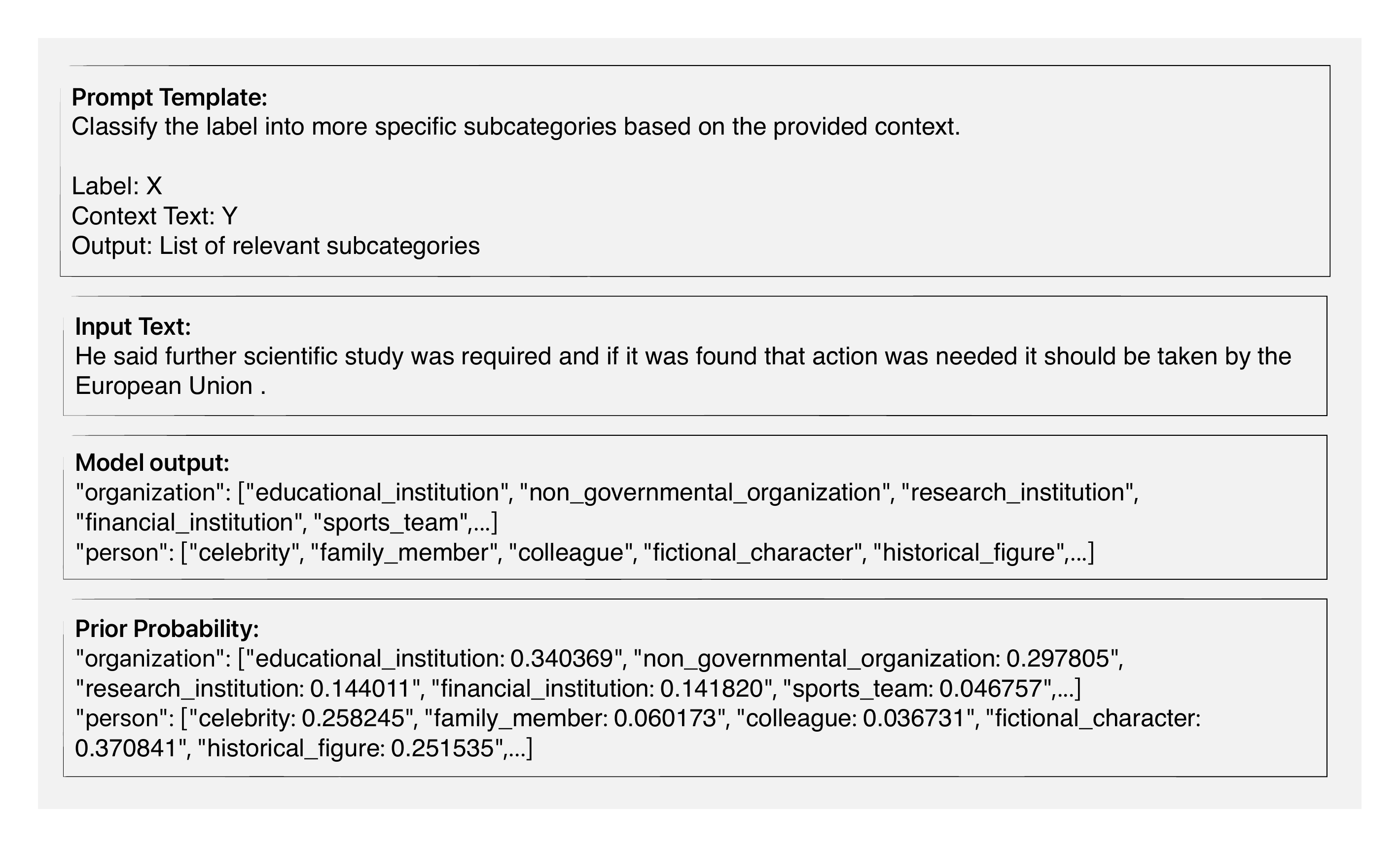}
\caption{Case study demonstrating the resampling stage with particle mutation. }
\label{fig:case3}
\end{figure*}
The final stage employs an LLM-based resampling mechanism that performs context-aware particle mutation. As shown in Figure\ref{fig:case3}, given the input text "He said further scientific study was required and if it was found that action was needed it should be taken by the European Union," the system generates contextually relevant subcategories.
The resampling process works as follows: the LLM analyzes the specific context and mutates the original particles to better fit the observed text patterns. For the "organization" label, the system generates context-specific subcategories such as "educational institution," "non-governmental organization," "research institution," and "financial institution," which are more relevant to the scientific and policy context of the input.
Prior probabilities are computed to indicate higher prior probability. For instance, "educational institution: 0.340369" and "non-governmental organization: 0.297805" receive relatively high priors due to their contextual relevance. This perplexity-based weighting ensures that contextually appropriate mutations are favored during the resampling process.

\section{Ablation on Semantic Decomposition}
\label{app:semantic_ablation}
To further validate the necessity of semantically coherent subcategories, 
we design a controlled ablation experiment that breaks the semantic alignment 
between subcategories and their corresponding entity labels.

We take the optimized subcategory set learned by BCL and randomly reassign 
subcategories to different entity labels. For example, subcategories such as 
``company'' or ``event'' may be assigned to the ``Person'' label instead of 
semantically consistent categories like ``athlete'' or ``politician''. 
Importantly, the subcategory pool remains unchanged, ensuring that the only 
difference lies in the loss of semantic alignment.

Table~\ref{tab:semantic_ablation} shows the results across representative 
datasets and models. Breaking semantic coherence leads to consistent 
performance degradation across all settings.

\begin{table}[h]
\centering
\small
\begin{tabular}{l|l|c|c|c}
\hline
Dataset & Model & Semantic & Shuffled & $\Delta$ \\
\hline
CoNLL03 & Qwen-7B & 72.83 & 67.20 & -5.63 \\
CoNLL03 & Llama-8B & 69.14 & 56.90 & -12.24 \\
ACE05 & Qwen-7B & 46.94 & 38.90 & -8.04 \\
ACE05 & Llama-8B & 53.10 & 49.50 & -3.60 \\
\hline
\end{tabular}
\caption{Effect of semantic decomposition.}
\label{tab:semantic_ablation}
\end{table}

On average, shuffling subcategories results in a 7.38 F1 point drop, 
demonstrating that semantic coherence plays a crucial role in enabling 
effective optimization. Meanwhile, the framework still maintains 
reasonable performance, suggesting that BCL exhibits graceful 
degradation rather than catastrophic failure.

\section{Frequency-Stratified Analysis on Long-Tail Entities}
\label{app:long_tail}

To further investigate the potential long-tail effect discussed in Section~X, 
we conduct a frequency-stratified evaluation to quantify model performance 
across entities with different training frequencies.

We perform experiments on two models (Llama-3.1-8B and Qwen2.5-7B) and two 
datasets (ACE05 and CoNLL2003). Entities in the test set are grouped into 
three tiers based on their frequency in the training set: Top 30\% (frequent), 
Mid 30\% (medium-frequency), and Bottom 40\% (rare). We report F1 scores for 
each group separately.

Table~\ref{tab:long_tail} summarizes the results. Overall, BCL demonstrates 
strong robustness to frequency imbalance. The performance gap between frequent 
and rare entities is generally small (within 1–2 F1 points), and in some cases, 
rare entities even achieve comparable or slightly better performance.

\begin{table*}[t]
\centering
\small
\begin{tabular}{l|l|c|c|c}
\hline
Model & Dataset & Top 30\% & Mid 30\% & Bottom 40\% \\
\hline
Llama-3.1-8B & ACE05 & 53.89 & 55.25 & 51.94 \\
Llama-3.1-8B & CoNLL2003 & 69.43 & 69.64 & 68.23 \\
Qwen2.5-7B & ACE05 & 46.62 & 47.61 & 47.03 \\
Qwen2.5-7B & CoNLL2003 & 72.39 & 73.31 & 73.67 \\
\hline
\end{tabular}
\caption{Frequency-stratified F1 performance of BCL.}
\label{tab:long_tail}
\end{table*}

These results suggest that, although the particle filtering mechanism 
theoretically emphasizes frequent patterns, its impact on rare entity 
performance is limited in practice. We hypothesize that the Bayesian 
aggregation process helps mitigate overfitting to frequent patterns by 
maintaining diverse hypothesis particles.

\section{Implementation Details}

\subsection{Prior and Likelihood Selection in Bayesian Filtering}
\label{app:weight}

In our Bayesian filtering framework, particle weights are updated through the 
combination of prior beliefs and observed evidence. We use perplexity-based 
scores as the prior and task-specific confidence as the likelihood, based on 
the following considerations:

\textbf{(1) Rationality of the Prior:} 
Perplexity reflects the degree of consistency between generated rules and the 
LLM's internal language distribution. Lower perplexity indicates that the rule 
better aligns with the model's ``linguistic intuition,'' and as a prior 
assumption, such rules are more likely to yield high-quality extraction results. 
This provides a reasonable inductive bias: before observing any task-specific 
performance, we favor rules that are linguistically coherent and natural 
according to the model's pre-trained knowledge.

\textbf{(2) Rationality of the Likelihood:} 
The IE confidence score $\text{L}_{\text{c}}$ (Equation~\ref{eq:confidence_definition}) 
measures how well a rule performs on actual extraction tasks. Unlike perplexity, 
which only captures linguistic properties, the confidence score directly evaluates 
the rule's effectiveness in guiding the LLM to generate correct entity labels. 
This task-specific measurement serves as the likelihood function $p(D|\theta)$, 
providing empirical evidence to update our beliefs about rule quality based on 
observed extraction performance.

\textbf{(3) Separation of Prior and Likelihood:} 
In standard Bayesian updating, the prior $p(\theta)$ and likelihood $p(D|\theta)$ 
serve distinct roles. Similarly, in our framework:
\begin{itemize}[itemsep=2pt]
    \item \textbf{Perplexity (Prior):} Encodes the inductive bias that 
    ``rules should be linguistically fluent and coherent''
    \item \textbf{IE Confidence (Likelihood):} Provides observational evidence 
    through actual task performance
\end{itemize}

This separation allows newly generated particles to carry reasonable initial 
beliefs into the evaluation stage, rather than using uninformative uniform priors. 
The Bayesian update mechanism (Equation~\ref{eq:update_unnorm}) then combines 
these two sources of information:
\begin{equation}
\underbrace{w_i^{(t)}}_{\text{posterior}} \propto 
\underbrace{w_i^{(t-1)}}_{\text{prior}} \cdot 
\underbrace{\exp(\beta \cdot y_i^{(t)})}_{\text{likelihood}}
\end{equation}

\textbf{(4) Robustness via Filtering Dynamics:}
Even if perplexity imperfectly approximates rule quality, the particle filtering 
framework provides inherent robustness through its denoising mechanism. Particles 
with misleading priors (low perplexity but poor IE performance) receive low 
likelihood scores, causing their weights to exponentially decay through 
multiplicative updates and be eliminated during resampling. This mechanism 
ensures that the final particle distribution is dominated by task performance 
rather than initial priors.

    

\subsection{Evaluation Metric.} 
\label{app:evaluation}
Following previous work~\citep{huang2025guidener}, 
we adopt entity-level F1 scoring for both NER and RE tasks. For NER, a predicted 
entity is considered correct only when both its boundary (character span) and 
type label exactly match the ground truth annotation. For RE, both entity mentions 
and their relation type must match exactly. Precision, recall, and F1 score are 
computed at the entity/relation level across the entire test set.

\subsection{Computational Complexity.} 
\label{app:complexity}
Let $K$ denote the number of particles per 
label and $M$ denote the number of training samples used for optimization. Our 
method processes batches sequentially through iterative filtering, requiring 
$O(K \times M)$ LLM inference calls in total. In contrast, methods like 
GuideNER~\citep{huang2025guidener} require $O(N)$ calls to traverse the complete 
training set of size $N$. Since we operate on $M \ll N$ samples (e.g., $M \approx 
0.03N$ as shown in Section~\ref{main:quantity}), our method achieves significant efficiency 
gains. The weight update (Equation~\ref{eq:update_norm}) and resampling 
(Equation~\ref{eq:resample_weights}) operations involve only $O(K)$ numerical 
computations---probability calculations, softmax normalization, sorting, and 
random sampling---which are negligible compared to LLM inference time. Therefore, 
the computational cost is dominated by LLM inference, and our approach is 
substantially more efficient than full-dataset traversal methods.

\subsection{Hyperparameters.}
\label{app:hyperparameters}
Key hyperparameters are set as follows: number of 
particles per label $N=10$ (empirically optimal range: 1--20), selection pressure 
$\beta=2.0$ (selected via grid search over $\{1, 2, 5, 10\}$ across multiple 
datasets), observation batch size of 4 sentences (analyzed in Section~\ref{app:context_length}), 
and retention ratio of 50\% for resampling (balancing exploitation and exploration). 
All LLM inference is performed with temperature=0.0 and random seed=42 for 
reproducibility. Convergence is reached when validation F1 score plateaus for 
3 consecutive iterations.

\section{Check List}
\label{sec:appendix_d}

This appendix provides additional details required by the ACL Responsible NLP Research checklist.

\subsection{Data Licenses and Terms of Use (B2)}
\label{sec:data_licenses}

We provide license information for all datasets used in our experiments:

\begin{table}[h]
\centering
\small
\begin{tabular}{lll}
\toprule
\textbf{Dataset} & \textbf{License/Terms} & \textbf{Access} \\
\midrule
\multicolumn{3}{l}{\textit{Named Entity Recognition}} \\
CoNLL-2003 & Research use only & Public \\
ACE 2005 & LDC User Agreement & LDC \\
GENIA & GENIA Project License & Public \\
\midrule
\multicolumn{3}{l}{\textit{Relation Extraction}} \\
NYT & LDC (derived) & Public \\
CoNLL04 & Research use only & Public \\
SciERC & CC BY 4.0 & Public \\
\bottomrule
\end{tabular}
\caption{License information for all datasets used in experiments.}
\label{tab:licenses}
\end{table}

\paragraph{Model Licenses.} 
The language models used in our experiments are distributed under the following licenses:
\begin{itemize}
    \item \textbf{Qwen2.5-3B/7B}: Apache 2.0 License (Qwen Team, Alibaba)
    \item \textbf{Llama-3.1-8B}: Llama 3.1 Community License (Meta)
    \item \textbf{Pixtral-12B}: Apache 2.0 License (Mistral AI)
\end{itemize}

\paragraph{Code Availability.}
Our implementation will be released under the MIT License upon acceptance. The codebase includes all scripts for data preprocessing, particle filtering optimization, and evaluation.

\subsection{Intended Use and Consistency (B3)}
\label{sec:intended_use}

All datasets were used in accordance with their intended purposes:

\begin{itemize}
    \item \textbf{CoNLL-2003}: Originally created for the CoNLL-2003 shared task on language-independent named entity recognition. We use it for NER evaluation as intended.
    
    \item \textbf{ACE 2005}: Developed for entity, relation, and event extraction research. We use the entity annotations for NER evaluation.
    
    \item \textbf{GENIA}: Created for biomedical text mining research. We use it for biomedical NER evaluation as intended.
    
    \item \textbf{NYT}: Derived from New York Times articles for relation extraction research. We use it for RE evaluation.
    
    \item \textbf{CoNLL04}: Designed for joint entity and relation extraction. We use it for RE evaluation.
    
    \item \textbf{SciERC}: Created for information extraction from scientific papers. We use it for scientific RE evaluation as intended.
\end{itemize}

Our derived artifacts are intended solely for research purposes in information extraction and should not be used for commercial applications without appropriate licensing.

\subsection{Privacy and Offensive Content (B4)}
\label{sec:privacy}
We did not perform additional anonymization as the original dataset creators have already addressed privacy considerations in their data collection and release procedures.

\subsection{Artifact Documentation (B5)}
\label{sec:artifact_doc}

\paragraph{Dataset Coverage.}
Table~\ref{tab:dataset_coverage} provides detailed documentation of the datasets used:

\begin{table}[h]
\centering
\small
\begin{tabular}{llll}
\toprule
\textbf{Dataset} & \textbf{Domain} & \textbf{Language} & \textbf{Text Source} \\
\midrule
CoNLL-2003 & News & English & Reuters \\
ACE 2005 & News/Conv. & English & Various news \\
GENIA & Biomedical & English & PubMed abstracts \\
NYT & News & English & New York Times \\
CoNLL04 & General & English & News articles \\
SciERC & Scientific & English & AI paper abstracts \\
\bottomrule
\end{tabular}
\caption{Domain, language, and source documentation for all datasets.}
\label{tab:dataset_coverage}
\end{table}

\paragraph{Entity and Relation Types.}
\begin{itemize}
    \item \textbf{CoNLL-2003 (4 types)}: PER, LOC, ORG, MISC
    \item \textbf{ACE 2005 (7 types)}: Person, Organization, Location, Facility, Weapon, Vehicle, GPE
    \item \textbf{GENIA (5 types)}: DNA, RNA, Protein, Cell Line, Cell Type
    \item \textbf{NYT (24 relation types)}: Including location-related, person-related, and organization-related relations
    \item \textbf{CoNLL04 (5 relation types)}: Located-In, Work-For, OrgBased-In, Live-In, Kill
    \item \textbf{SciERC (7 relation types)}: Used-for, Feature-of, Part-of, Compare, Hyponym-of, Evaluate-for, Conjunction
\end{itemize}

\subsection{Package Parameters (C4)}
\label{sec:package_params}

\paragraph{Software Dependencies.}
\begin{itemize}
    \item Python 3.10.12
    \item PyTorch 2.6.0
    \item Transformers 4.51.3
    \item NumPy 1.24.3
    \item scikit-learn 1.3.0 (for evaluation metrics)
\end{itemize}

\paragraph{Evaluation Implementation.}
We use the \texttt{seqeval} library (v1.2.2) for NER evaluation with the following settings:
\begin{itemize}
    \item \texttt{mode='strict'}: Exact boundary and type matching required
    \item \texttt{scheme=IOB2}: Using IOB2 tagging scheme
\end{itemize}

For RE evaluation, we implement custom evaluation following prior work \cite{riedel2010modeling}:
\begin{itemize}
    \item A relation is correct if both entity mentions and relation type match exactly
    \item Precision, recall, and F1 are computed at the relation-tuple level
\end{itemize}

\paragraph{Text Preprocessing.}
\begin{itemize}
    \item Tokenization: Model-specific tokenizers from HuggingFace
    \item No additional preprocessing (lowercasing, stemming) applied
    \item Maximum sequence length: 512 tokens (truncation applied if exceeded)
\end{itemize}

\subsection{Use of AI Assistants (E1)}
\label{sec:ai_assistants}
In this paper, we employed Large Language Models (LLMs) as core components of our proposed BCL framework in three specific stages. First, we utilized LLMs for rule extraction and particle generation to create initial subcategory patterns from training data, ensuring systematic rule discovery. Second, we leveraged LLMs for performance evaluation through in-context learning inference, where models assess rule effectiveness on validation datasets. Third, we employed LLMs for rule mutation and resampling to generate semantically diverse rule variants during the optimization process. 

Additionally, we used LLMs to assist with writing and language improvement throughout the manuscript preparation process. This included grammar checking, sentence structure optimization, and clarity enhancement to improve the overall readability of our work. However, all core research ideas, methodological contributions, experimental designs, and conclusions remain entirely our own intellectual work.

\subsection{Broader Impact Statement}
\label{sec:broader_impact}

\paragraph{Positive Impacts.}
BCL advances information extraction technology with several benefits:
\begin{itemize}
    \item \textbf{Accessibility}: By improving ICL performance on smaller models (3B-12B parameters), BCL democratizes access to effective IE systems for researchers and practitioners with limited computational resources.
    \item \textbf{Efficiency}: The data-efficient optimization (converging with 3-5\% of training data) reduces the annotation burden and computational costs.
    \item \textbf{Generalizability}: The framework's applicability to both NER and RE tasks provides a unified approach to diverse IE challenges.
\end{itemize}

\paragraph{Potential Negative Impacts.}
\begin{itemize}
    \item \textbf{Automation of Information Extraction}: While beneficial for legitimate applications, improved IE could potentially be misused for unauthorized data harvesting or surveillance.
    \item \textbf{Environmental Impact}: Although more efficient than full fine-tuning, LLM-based IE still requires significant computational resources with associated carbon emissions.
\end{itemize}

\paragraph{Mitigation Strategies.}
We encourage users to:
\begin{enumerate}
    \item Apply BCL only to data they have legal rights to process
    \item Implement human oversight in high-stakes applications
    \item Consider the environmental impact and use appropriately-sized models for their needs
\end{enumerate}




\end{document}

%% file: math_commands.tex
\usepackage{amsmath,amsfonts,bm}

\def\eqref#1{equation~\ref{#1}}

\def\1{\bm{1}}

\DeclareMathAlphabet{\mathsfit}{\encodingdefault}{\sfdefault}{m}{sl}
\SetMathAlphabet{\mathsfit}{bold}{\encodingdefault}{\sfdefault}{bx}{n}

\DeclareMathOperator*{\argmax}{arg\,max}